\documentclass[fleqn,10pt,twocolumn]{ICCAS2019}
 
 \usepackage[dvipsnames]{xcolor}
 \usepackage{subfig}
 
\begin{document}

\title{Multi-lane Detection Using Instance Segmentation and Attentive Voting}

\author{Donghoon Chang${}^{1}$, Vinjohn Chirakkal${}^{2}$, Shubham Goswami${}^{1,*}$, Munawar Hasan${}^{1}$, Taekwon Jung${}^{2}$, Jinkeon Kang${}^{1,3}$, Seok-Cheol Kee${}^{4}$, Dongkyu Lee${}^{5}$, Ajit Pratap Singh${}^{1}$ }

\affils{ ${}^{1}$Department of Computer Science, IIIT-Delhi, India \\
${}^{2}$Springcloud Inc., Korea \\
${}^{3}$Center for Information Security Technologies (CIST), Korea University, Korea \\
${}^{4}$Smart Car Research Center, Chungbuk National University, Korea \\
${}^{5}$Department of Smart Car Engineering, Chungbuk National University, Korea \\ ${}^{*}$ Corresponding author}

\abstract{
Autonomous driving is becoming one of the leading industrial research areas. Therefore many automobile companies are coming up with semi to fully autonomous driving solutions. Among these solutions, lane detection is one of the vital driver-assist features that play a crucial role in the decision-making process of the autonomous vehicle. A variety of solutions have been proposed to detect lanes on the road, which ranges from using hand-crafted features to the state-of-the-art end-to-end trainable deep learning architectures. Most of these architectures are trained in a traffic constrained environment. In this paper, we propose a novel solution to multi-lane detection, which outperforms state of the art methods in terms of both accuracy and speed. To achieve this, we also offer a dataset with a more intuitive labeling scheme as compared to other benchmark datasets. Using our approach, we are able to obtain a lane segmentation accuracy of $99.87\%$ running at 54.53 fps (average).
}

\keywords{
Autonomous Driving, Lane Detection, Segmentation, Deep Learning, Computer Vision.
}
\maketitle

\section{Introduction}

In recent years, self-driving cars have started becoming a reality due to the advances in deep learning and hardware capabilities of various sensors and control modules. It is seeking the attention of both industry and academic research as it covers a wide range of research areas from computer vision to signal processing and many more. Multi-lane detection is one of the vital driver-assist features in these autonomous vehicles, which plays a crucial role in path planning and the decision-making process of autonomous vehicles. It introduces several challenges and requires to be highly accurate and to work in real-time.

\subsection{Related Work}

A variety of lane detection methods have been proposed starting from the traditional methods using hand-crafted features (\cite{ref1}, \cite{ref2}, \cite{ref3}, \cite{ref4}, \cite{ref5}, \cite{ref6}) to the modern state of the art end-to-end trainable deep architectures (\cite{ref7}, \cite{ref8}, \cite{ref9}, \cite{ref10}, \cite{spcnn}, \cite{te2e}). Spatial CNN \cite{spcnn} and end-to-end lane segmentation \cite{te2e} use CNN-based approach to exploit the spatial information in a road scene and try to develop an understanding of a road scene. The key issue with these architectures is that they are not trained in adverse conditions, and the dataset used for training comes from traffic constrained environment and also has a non-intuitive annotation procedure as it is discussed further in Section \ref{sec.2}.

\subsection{Motivation}

Semantic segmentation of a road scene involves a variety of objects with different sizes. The encoders in segmentation network architectures like \cite{segnet}, \cite{unet}, and \cite{enet} are deep enough that neurons in its last layer have a receptive field greater than or equal to the size of the input images. In lane segmentation, lane markings are placed on a road; hence, there is an upper bound on the size of lane markings. It can be exploited by limiting the depth of the encoder, by increasing the inference speed and by decreasing learnable parameters in lane segmentation network. Apart from lane segmentation, the state-of-the-art method like \cite{te2e} strongly depends on the road scene environment to generate the embeddings for clustering. Intuitively, using lane markings alone for multi-lane detection makes more sense. This also reduces the need to adequately capture the background diversity and dependencies, such as traffic and geometric constraints. For example, VPGNet \cite{samsungr} predicts vanishing point on the input frame for guiding lane and road marking detection and recognition in adverse circumstances this introduces more training data to capture the input diversity for vanishing point prediction and also vanishing point works well on straight roads but not on curving roads or roundabouts.

\section{Dataset} \label{sec.2}

\begin{figure}[t]
\begin{center}
\includegraphics[width=8.0cm]{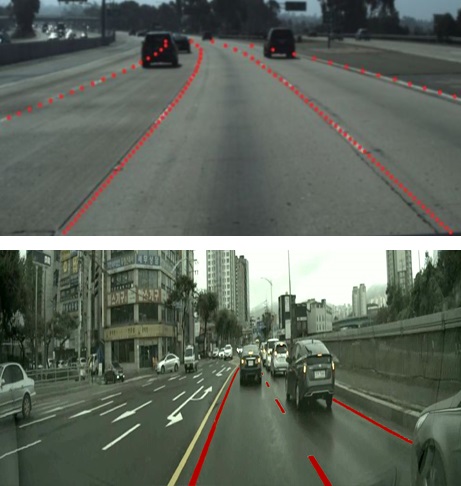}
\caption{Top: an annotated sample frame from tuSimple dataset. Bottom: an annotated sample frame from our dataset.}
\label{comparison}
\end{center}
\end{figure}

There are various available road scene datasets with pixel-level annotations; CamVid (Cambridge-driving Labeled Video Database, \cite{camvid}), Cityscapes \cite{citysc}, KITTI dataset \cite{kitti}, Caltech lane dataset \cite{caltechlane} and tuSimple \cite{tusimple}. Apart from tuSimple, others contain a small set of images with lane marking annotations. On the other hand, datasets like tuSimple are very traffic-constrained (light traffic and clear lane markings in \cite{spcnn}) and its annotations are comprised of continuous lane curves starting from the bottom of the input frame till the horizon and even passing over the vehicles as shown in Fig. \ref{comparison} (Top). The issue with such labeling approach is that only lane marking pixels should be marked as lane markings, and one shall not assume them anywhere else; otherwise, this can cause issues by predicting false positives in high traffic scenarios \cite{spcnn}.

We collected a small dataset which includes city traffic on cloudy days capturing conditions ranging from water reflection on the road to moving wipers on the windshield. From this collected dataset, we annotated $5,000$ frames in a pixel-wise manner without carrying any assumptions, i.e., only visible lane markings aligned in the direction of the movement of the vehicle, as shown in Fig. \ref{comparison} (Bottom).

\section{Methodology}

\begin{figure}[thb]
\begin{center}
\includegraphics[width=8cm]{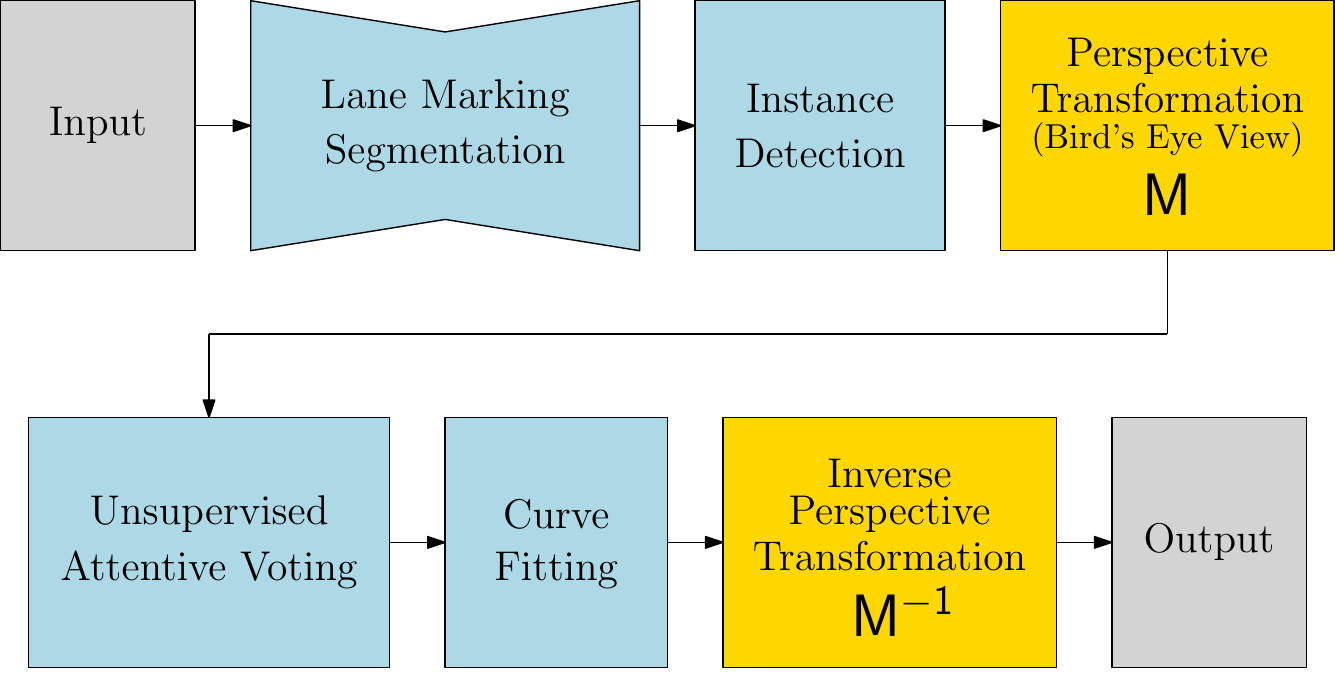}
\caption{Architecture Diagram}
\label{fig:architecture}
\end{center}
\end{figure}

The preprocessing of input frames involves cropping majority of the sky portion and car dashboard. Afterward, the frame is resized to the resolution of $360$x$480$. This frame is then fed to the lane marking segmentation network, which segments out the visible lane marking pixels followed by detecting instances of segmented lane markings using graph-based methods. Perspective transformation (bird's eye view) is then applied to the instance segmented output followed by attentive voting based clustering method and polynomial curve fitting, which provides the final output. The architecture diagram is given in Fig.\ref{fig:architecture}.

\subsection{Lane segmentation}

We follow CNN-based approach \cite{imagenet} to create lane segmentation network with \textit{strided convolutions} and \textit{strided de-convolutions} with \textit{relu} \cite{relu} activation in hidden units and we propose the customized soft dice loss \cite{diceloss} to penalize false positives by modifying the ground truth (Eq. (\ref{eq.1})) in the numerator of $Loss_{custom}$ (Eq. (\ref{eq.3})). The strided convolution is used to increase the receptive field.

\begin{figure}[thb]
\begin{center}
\includegraphics[width=8.2cm]{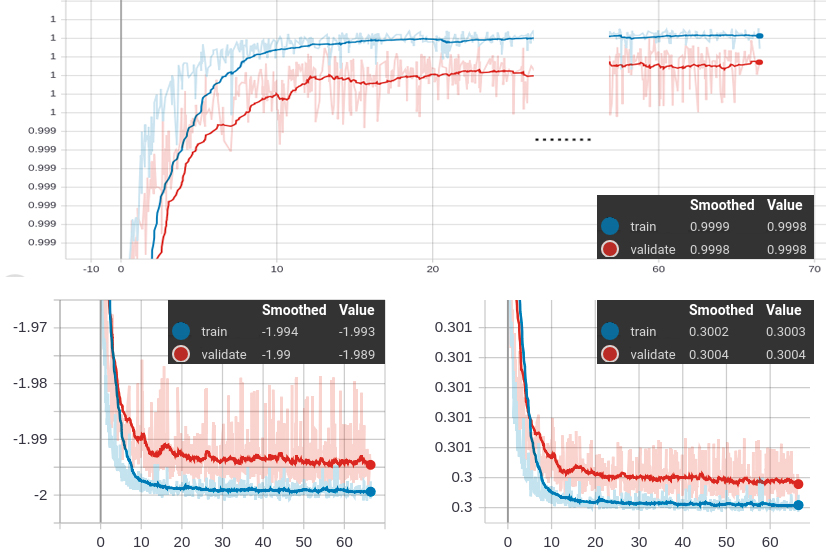}
\caption{Top: accuracy plot. Bottom: $Loss_{custom}$ (left) and softmax loss (right).}
\label{acc_loss}
\end{center}
\end{figure}

Fig. \ref{acc_loss} shows the accuracy, loss plot and softmax loss used as the evaluation metrics. Table \ref{network} shows the lane segmentation network architecture. We have exploited the upper bound on the size of lane marking and the last layer of encoder captures a receptive field of $63$x$63$ $pixel^{2}$ on the original input.

\begin{align}
&\hat{y_{gt}}[i][j][k] = -\alpha~~where~y_{gt}[i][j][k] = 0 \label{eq.1} \\
&mean(A) = \frac{\sum_{i}\sum_{j}A_{i,j}}{\|A\|} \label{eq.2}\\
\begin{split}
&Loss_{custom}(y_{gt}, y_{pred}) = \\ 
&- \sum_{k=1}^{c}\frac{2*mean(\hat{y_{gt}}^{k}\circ y_{pred}^{k}) + \epsilon}{mean(y_{gt}^{k}\circ y_{gt}^{k})+mean(y_{pred}^{k}\circ y_{pred}^{k}) +\epsilon}
\label{eq.3}
\end{split}
\end{align}

In Eq. (\ref{eq.1}), $i$ and $j$ represent the spatial index. $k$ represents the channel index of the output, where $k=1$ and $k=2$ are the channel for background (non-lane marking) and lane marking, respectively. In Eq. (\ref{eq.3}), $c$ is the total number of channels in the ground truth/prediction, $y_{pred}^{k}$, $y_{gt}^{k}$ and $\hat{y_{gt}}^{k}$ are the $k$-th channel values of predicted output, the ground truth and modified ground truth (Eq. (\ref{eq.1})), respectively. We achieved the best results with $\alpha=10^{-2}$ and $\epsilon=10^{-5}$ after tuning them respectively while training.

\begin{center}
\renewcommand{\arraystretch}{1.4}
\begin{table}[htb]
\caption{Lane segmentation network architecture with fused batch normalization \protect\cite{fusedbtn} }
\begin{tabular}{ | m{9em} m{5em} m{5em} | } 
\hline Layer & Input & Output\\
\hline
Conv2D+Relu. Conv2D+Relu. Conv2D+Relu. Conv2D+Relu. Conv2D+Relu. & 360x480x3 179x239x32 89x119x64 44x59x128 21x29x256& 179x239x32 89x119x64 44x59x128 21x29x256 10x14x512\\ 
\hline
Conv2DTrans.+Relu. Conv2DTrans.+Relu. Conv2DTrans.+Relu. Conv2DTrans.+Relu. Conv2DTrans.& 10x14x512 21x29x256 44x59x128 89x119x64 179x239x32& 21x29x256 44x59x128 89x119x64 179x239x32 360x480x2\\ 
\hline
Softmax & 360x480x2 & 360x480x2 \\ 
\hline
\end{tabular}
\label{network}
\end{table}
\end{center}

\subsection{Instance detection and bird's eye view}

We use the customized Breadth-first search to detect all the distinct instances of lane markings in the output of the lane segmentation network. Perspective transformation is then applied to provide ease of voting in clustering and curve fitting as the bird's eye view reduces the degree of a polynomial. It is important because on many occasions only a few pixels are obtained for some lane markings (e.g., lane markings on road boundaries) and higher degree polynomial requires proportionally more pixels to fit; thus a lower degree polynomial helps in performing the better curve fitting even if the instance pixels are less in number.

\subsection{Unsupervised attentive voting and curve fitting}

\begin{figure}[t]
\begin{center}
\includegraphics[width=6.6cm]{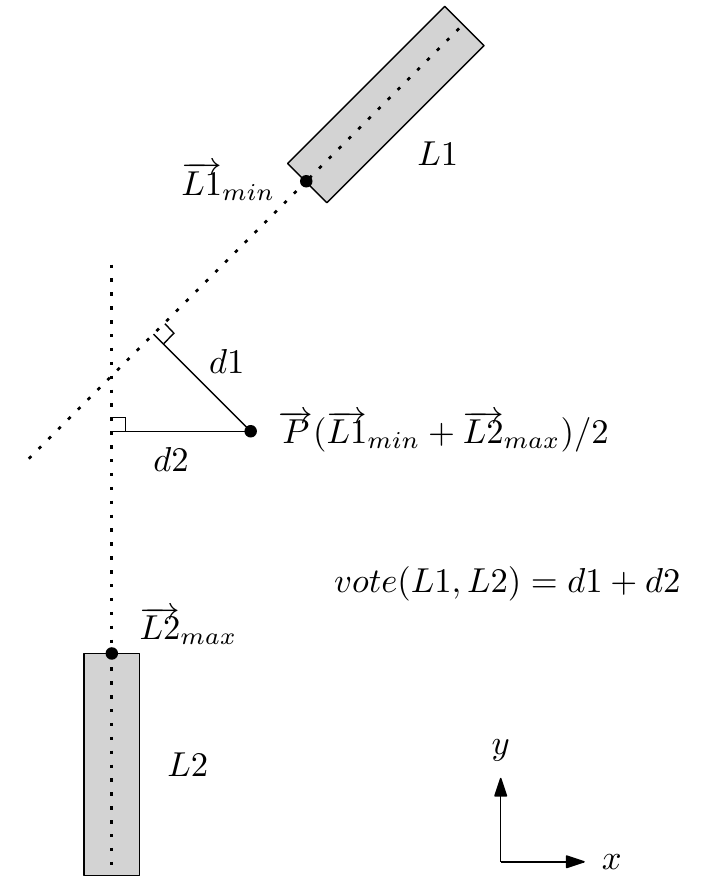}
\caption{The example of attentive voting between two instances of lane markings $L1$ and $L2$.}
\label{voting}
\end{center}
\end{figure}

In the bird's eye view space, all the instances vote for their closest instance based on slope and spatial positioning of the instances. The slope and spatial positioning of an instance (i.e., lane marking) are used as spatial attention to vote for its closest lane marking, which belongs to the same lane divider. Fig. \ref{voting} shows how the attention is used to calculate the vote by giving the example of attentive voting among a pair of lane marking instances. $Li$ represents the instance of lane marking. $\overrightarrow{Li}_{min}$ and $\overrightarrow{Li}_{max}$ are the bottom-most and the top-most pixel of lane marking $Li$, respectively. The vote between two lane makings $Li$ and $Lj$ as $vote(Li,Lj)$ which is calculated by taking sum of $d1$ and $d2$ where $di$ (for $i \in \{1,2\}$ in Fig. \ref{voting}) represents the perpendicular distance of point $P$ with the line fitted on the pixels of lane marking $L_i$ using normal equation (closed-form solution for linear regression \cite{bishop}). If this vote is less than some threshold $\eta$, then these two instances belong to the same lane divider. This unsupervised clustering approach takes care of some misclassified lane marking pixels as through attentive voting the instance belonging to same lane divider are correctly grouped together, which is the expected outcome. Curve fitting for a second-degree polynomial is performed in the same bird's eye view space on each cluster of instances and projecting this output back to original space provides the desired result.

\begin{figure*}
\centering
\subfloat{\includegraphics[width=0.196 \textwidth]{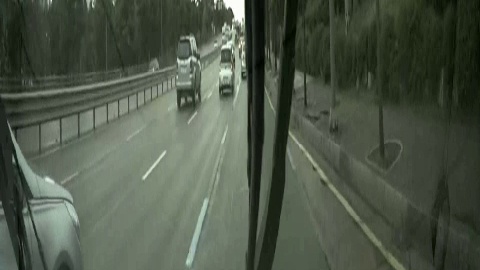}} 
\subfloat{\includegraphics[width=0.196\textwidth]{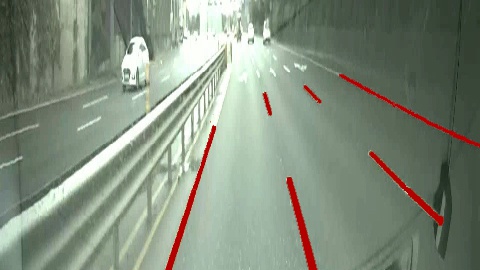}} 
\subfloat{\includegraphics[width=0.196\textwidth]{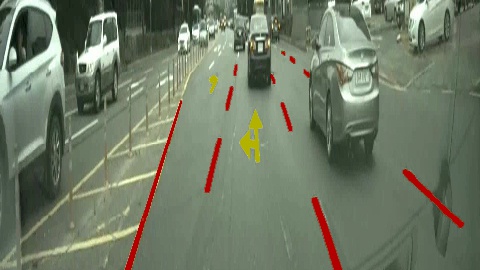}}
\subfloat{\includegraphics[width=0.196\textwidth]{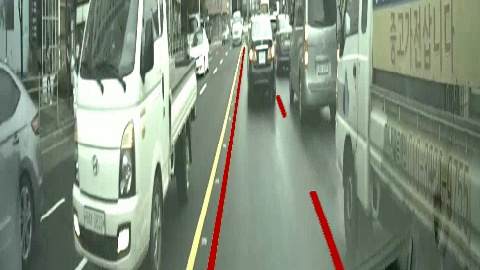}}
\subfloat{\includegraphics[width=0.196\textwidth]{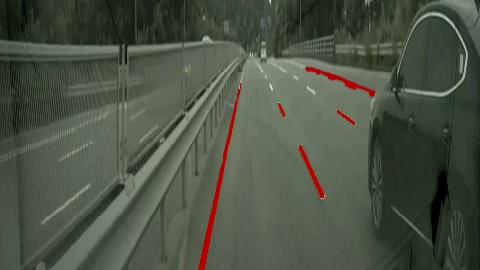}}\\
\subfloat{\includegraphics[width=0.196\textwidth]{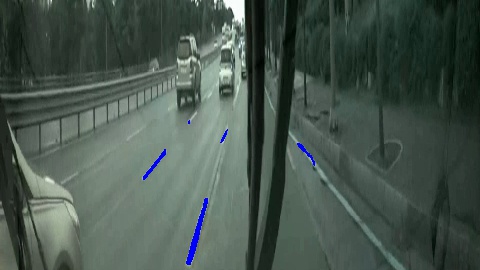}}
\subfloat{\includegraphics[width=0.196\textwidth]{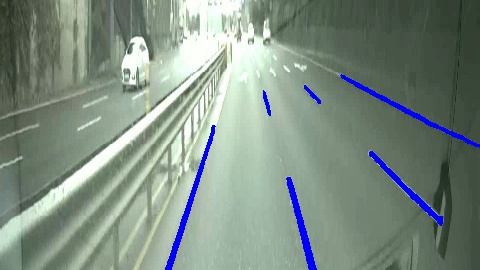}}
\subfloat{\includegraphics[width=0.196\textwidth]{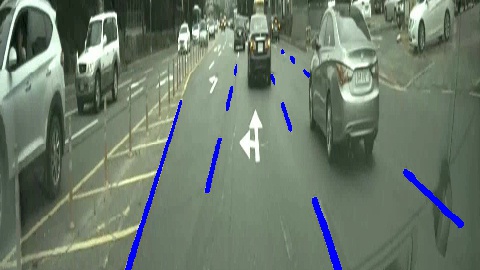}} 
\subfloat{\includegraphics[width=0.196\textwidth]{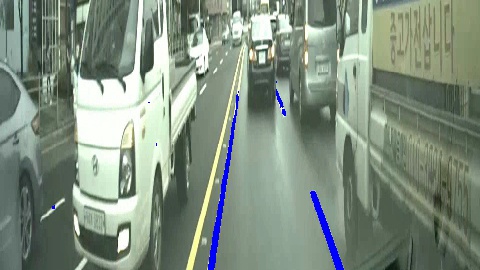}}
\subfloat{\includegraphics[width=0.196\textwidth]{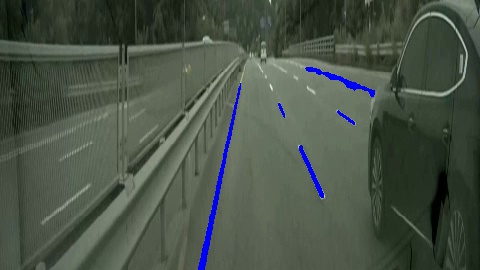}}\\
\subfloat{\includegraphics[width=0.196\textwidth]{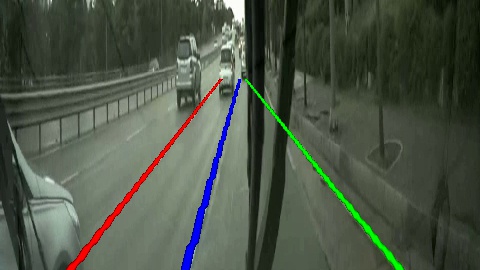}} 
\subfloat{\includegraphics[width=0.196\textwidth]{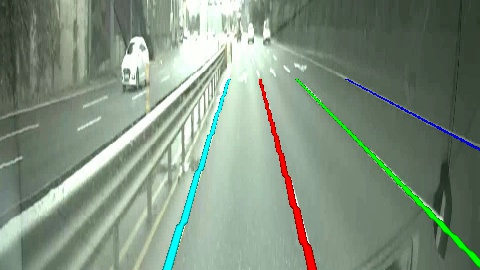}}
\subfloat{\includegraphics[width=0.196\textwidth]{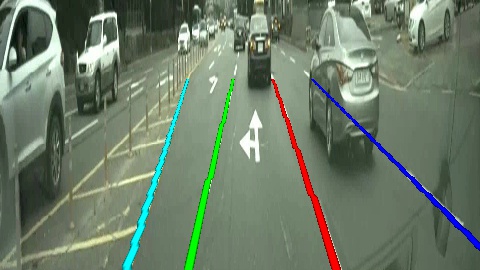}}
\subfloat{\includegraphics[width=0.196\textwidth]{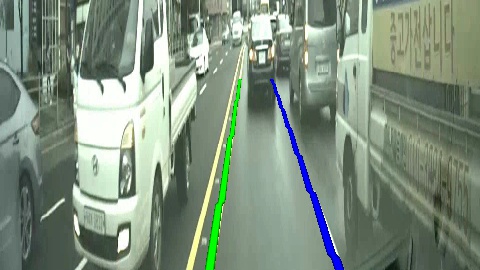}}
\subfloat{\includegraphics[width=0.196\textwidth]{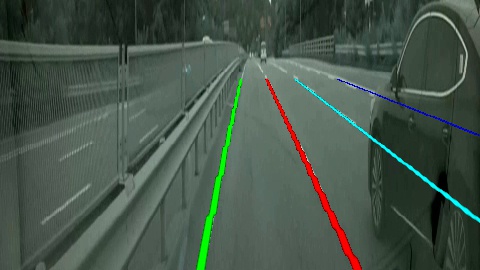}}\\
\caption{Visual result. Top row: original input with ground truth. Middle row: the output of lane segmentation network. Bottom row: the final output of attentive voting and curve fitting.}
\label{fig:det}
\end{figure*}

\section{Experiments and Results}

For our experiments, we ran our model on Intel Xeon(R) E5-2620 v4 processor and Nvidia RTX 2080Ti. We calculate the average inference time approximately, as shown in Table \ref{inference_time}. It shows that the total processing time takes about $0.01833$ ms. In Table \ref{comparision_accuracy}, we calculate the accuracy of lane segmentation for the comparison. Fig. \ref{fig:det} shows the visual results of the proposed architecture in three different stages; the original input with ground truth (top), the output of lane segmentation network (middle), and the final output of attentive voting and curve fitting (bottom).

\begin{table}[h!]
\setlength{\extrarowheight}{0.75ex}
\caption{Inference time (approximate average)}
\label{inference_time}
\begin{center}
\begin{tabu}to\linewidth{|X[0.5c]|X[0.3c]|X[0.3c]|}\hline
Task & Time (ms) & Speed (fps) \\\hline
Lane segmentation & 0.001449 & 690 \\\hline
Instance detection + Bird's eye view + Attentive voting + Curve fitting & 0.01688 & 59.21 \\\hline
Total & 0.01833 & 54.535 \\\hline
\end{tabu}
\end{center}
\end{table}

\begin{table}[h!]
\setlength{\extrarowheight}{0.75ex}
\caption{Comparison of the accuracy of lane segmentation}
\label{comparision_accuracy}
\begin{center}
\begin{tabu}to\linewidth{|X[0.3c]|X[0.3c]|X[0.3c]|}\hline
& Accuracy & Speed (fps) \\ \hline
\cite{te2e} & 96.4\% & 52.6 \\\hline
Ours & 99.87\% & 54.535 \\\hline
\end{tabu}
\end{center}
\end{table}

\section{Conclusion}

In this paper, we proposed a novel method for multi-lane detection, which outperforms the state-of-the-art methods in terms of accuracy and speed. To this, we also offer the dataset with a more intuitive labeling scheme as compared to other benchmark datasets. Using our approach, we are able to obtain the accuracy of $99.87\%$ at $54.53$ fps.

\section*{Acknowledgements}

This work was supported by Korea Institute for Advancement of Technology (KIAT) grant funded by the Korea Government (MOTIE) (N0002428, The Competency Development Program for Industry Specialist)

\end{document}